\documentclass{article}

\usepackage[final]{nips_2017}

\usepackage[utf8]{inputenc} 
\usepackage[T1]{fontenc}    
\usepackage{hyperref}       
\usepackage{url}            
\usepackage{booktabs}       
\usepackage{amsfonts}       
\usepackage{nicefrac}       
\usepackage{microtype}      
\usepackage[pdftex]{graphicx}
\usepackage{wrapfig}
\usepackage{subfigure}

\usepackage{amsmath}
\usepackage{algorithm}
\usepackage[noend]{algpseudocode}
\makeatletter
\def\BState{\State\hskip-\ALG@thistlm}
\makeatother

\title{Relating Input Concepts to Convolutional \\Neural Network Decisions}

\author{
  Ning Xie, Md Kamruzzaman Sarker, Derek Doran, Pascal Hitzler, Michael Raymer\\
  Department of Computer Science \& Engineering\\
  Wright State University, Dayton OH, USA\\
  \texttt{xie.25@wright.edu} \\
}

\begin{document}

\maketitle

\begin{abstract}
Many current methods to interpret convolutional neural
 networks (CNNs) use
visualization techniques and words to highlight
concepts of the input seemingly relevant to a CNN's decision. 
The methods hypothesize that the recognition of these concepts are instrumental in the 
decision a CNN reaches, but the nature of this relationship has not been well
explored. To address this gap, this paper examines the quality of a concept's recognition 
by a CNN and the degree to which the recognitions 
are associated with CNN decisions. The study considers a CNN trained for scene recognition over
the ADE20k dataset. It uses a novel approach to find and score the 
strength of minimally distributed representations of input concepts (defined by objects in scene images) across 
late stage feature maps. Subsequent analysis finds evidence that 
concept recognition impacts decision making. 
Strong recognition of concepts frequently-occurring in few scenes are indicative of 
correct decisions, but recognizing concepts common to many scenes may mislead the network. 
\end{abstract}
 
\section{Introduction}
\label{sec:intro_section}

CNNs are a mainstay model for classification  
in computer vision~\citep{lecun1998gradient,girshick2014rich,ren2015faster,simonyan2014very,sun2014deep}.
While their performance is impressive, CNNs are 
opaque or ``black box'' in nature, and there is a growing concern that the inability to interpret 
their internal actions will hinder human confidence and
trust of these systems in practice~\citep{lipton16, doran17}. 
A number of current efforts to make CNNs interpretable relates internal node activations to 
aspects of the input image. An aspect may be a particular color or texture pattern,
like those processed in early stage CNN feature maps. Aspects
may also be broad patterns that define objects (or object parts) 
depicted in an image. Semantically meaningful image aspects like
pointy ears, paws and whiskers may 
lead a human to decide that an image is of a cat, while observing 
sand, water, blue sky, and shells in an image may
determine that the image depicts a beach. We define a semantically meaningful image aspect to 
be an {\bf input concept}.

Most current research relates node activations to input concepts by visualization techniques.
For example, \citet{zeiler2010deconvolutional}
developed the idea of a {\em deconvolution} where activations across
feature maps can be related to patterns in an input 
image. More recently, \citet{selvaraju2016grad} developed coarse localization maps based on a broad pattern of the input image and the gradient in a CNN model to highlight the associated network regions.
\citet{dosovitskiy2016inverting} and \citet{mahendran2015understanding}, on the other hand, find `hidden' features used by a CNN via an inversion process with up-convolutional neural networks. 
\citet{zhang2016top} generates task-specific attention maps for the input image via excitation
backprop. 

While the aforementioned techniques provide nice viewpoints into how internal activations
may be related to qualities of an input, there has been few research into whether 
the input concepts recognized are associated with the decisions made by a CNN. 
\citet{zintgraf2017visualizing}, \citet{bach2015pixel}, and \citet{montavon2017explaining} developed ways to measure how every input pixel supports a CNN's classification result by conditional multivariate model, layer-wise relevance backpropagation method, and deep Taylor decomposition respectively. 
However, these methods focus on pixel-level explanation, 
it remains unclear if {\em groups} of pixels representing an input concept highlighted in the resulting visualizations have an impact on CNN decisions.

In this paper, we investigate the relationship between how well a CNN recognizes
input concepts from an image and the decisions it makes. 
We specifically consider input concepts and decisions under a scene recognition task over the ADE20k 
dataset~\citep{zhou2017scene}.
The study is powered by a novel algorithm to compute how well {\em any} concept is recognized 
across the feature maps of a convolutional layer. 
Analysis along concept types, including those that appear often within
a scene, often across multiple scenes, and those unique to a scene 
reveal a weak relationship between correct decision making
and concept recognition. This relationship is dampened by 
the recognition of `sparse' concepts that seldom appear in the images of a scene and by `misleading' concepts that
appear often across the images of many different scenes. However, the recognition of concepts that are unique 
to the images of specific scenes promote correct CNN decisions. 

\section{Concept recognition}
\label{sec:concept_recognition}
Studying the relationship between input concepts and CNN decisions requires a measure of how
well such concepts are recognized by a CNN. 
We define a concept as being `recognized' if there are a set of late stage convolutional layer nodes 
that only activate over the the input because of the concept's presence.
Whereas much of the research assumes that these nodes must lie within
the same CNN feature map~\citep{bau2017network,zintgraf2017visualizing}, we assert that 
concept recognition could occur in a {\em distributed way}, across many feature maps at a 
convolutional layer. Past studies have suggested and demonstrated that 
neural networks learn a representation of input features in a distributed 
fashion~\citep{carpenter1988art,bengio2003neural,hinton1986learning};
thus, we do not consider the possibility that input concepts can only be recognized within a single feature map. 

In the context of scene classification, the recognition of a 
concept (e.g. an annotated object) would be manifested by a set of (distributed) nodes 
(across multiple feature maps)
that collectively respond to the input pixels representing the concept. 
If the set of nodes is a ``good'' recognizer of the concept, they should
collectively respond to all pixels representing the concept, and over
no pixels not representing the concept. We call a node activated if it takes
on a non-zero value under a sigmoid or tanh non-linearity, or is $> 0$ under a ReLU non-linearity. 

\begin{figure}
\centering
\begin{minipage}{.42\textwidth}
  \centering
    \includegraphics[width=\textwidth]{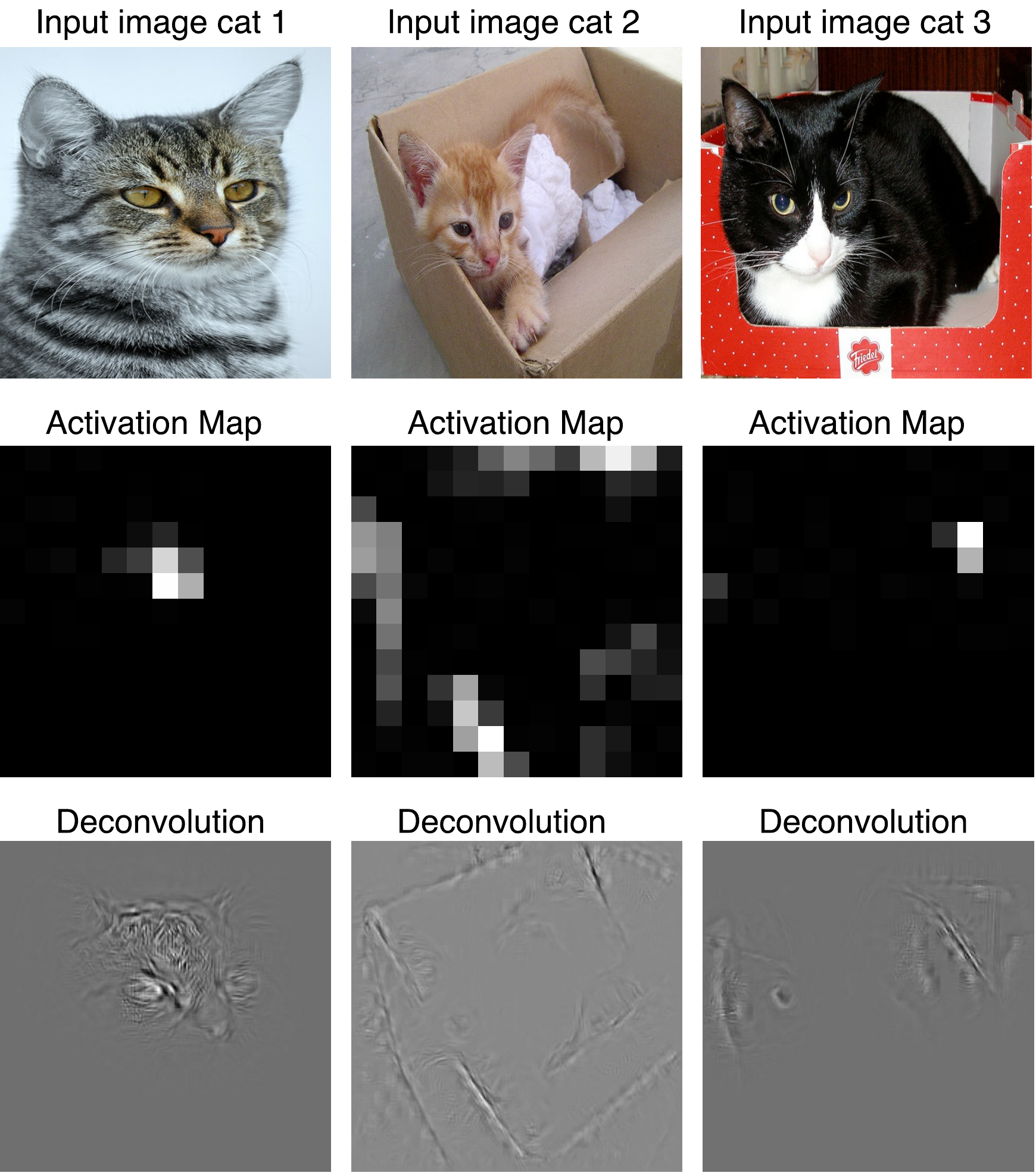}
 \caption{Deconvolutions of different cat images over the same feature map}
  \label{fig:nips-imagenet2}
\end{minipage}%
\hspace{20px}
\begin{minipage}{.52\textwidth}
  \centering
  \includegraphics[width=\textwidth]{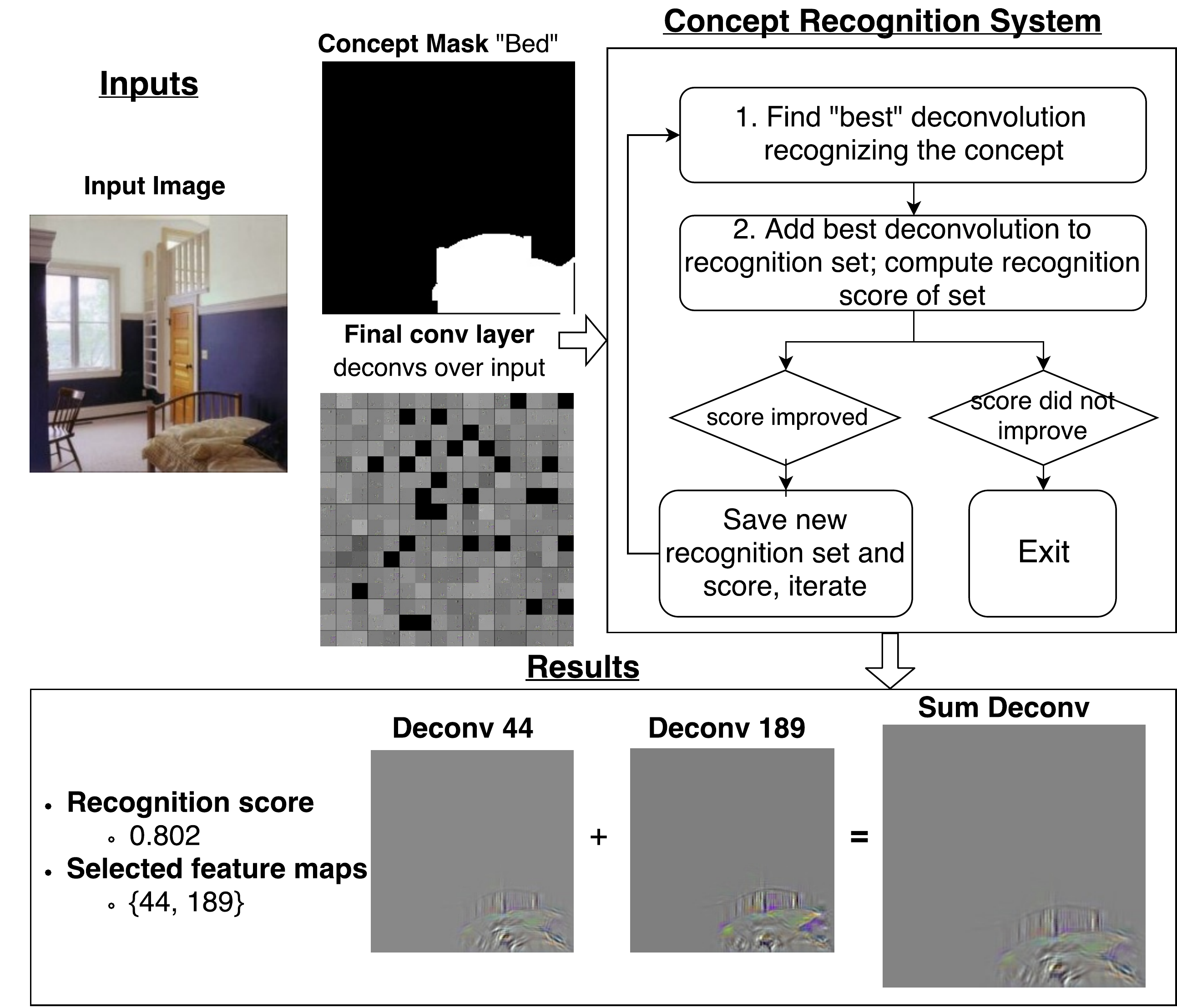}
\caption{Concept recognition across feature maps}
\label{fig:nips-recovery-framework}
\end{minipage}
\end{figure}

The deconvolution of a feature map recovers the pixels of an input image causing
its nodes to activate~\citep{zeiler2014visualizing,zeiler2011adaptive,yosinski2015understanding}. 
Deconvolutions thus seem like a natural way to identify if input concepts in scenes 
are represented by a feature map: if the deconvolution of the feature map covers most pixels of a concept, 
we may consider it as `recognized' by the feature map. However, patterns activating nodes in a feature
map are not always consistent from image to image. 
We illustrate this point in Figure~\ref{fig:nips-imagenet2} where
a feature map, taken from the last convolutional layer of AlexNet trained for object recognition, 
has its deconvolution computed for different input images. The deconvolution over the first
cat image suggests that the feature map recognizes
the facial features of a cat, or the
texture of a cat's fur. The deconvolution over the
second image, however, recognizes nothing about the
cat, and it is unclear if any concept in the
third image is recognized by the feature map. Recent 
approaches for concept recognition 
find that only a limited number of feature maps consistently 
recognize a specific concept~\citep{bau2017network}.

Instead of focusing on concept recognitions localized to a single feature map, 
Figure~\ref{fig:nips-recovery-framework} summarizes our approach to find and evaluate concepts recognized
{\em across} multiple feature maps in a convolutional layer. Given a binary segmentation mask of the concept and the deconvolutions of 
feature maps in the latest stage convolutional layer, a greedy algorithm selects the subset of feature
maps that collectively ``best" recognize the given concept according to a scoring function. 
The selected feature maps and a recognition quality score is then returned to the user. 
The specifics of the recognition scoring and the greedy algorithm are discussed next.

\subsection{Recognition scoring}
\label{subsub:score}
Ideally, the pixel area for a given concept should be covered by the deconvolutions of the selected feature maps as precisely as possible. The score should thus consider the combined coverage of the 
deconvolutions of the chosen feature
maps over and not over the pixels of a concept.  Based on this idea,
we evaluate how well a set of feature maps $G_\mathfrak{c}$ recognizes a concept $\mathfrak{c}$ in an image $\xi$ 
using a binary segmentation mask $M_\mathfrak{c}(\xi)$ that denotes the pixel positions of $\mathfrak{c}$ in $\xi$.
We assume that $M_\mathfrak{c}(\xi)$ is available in a dataset or can be generated via object segmentation
methods~\citep{chen2016deeplab}. 
From the set of deconvolutions $D_\mathfrak{c}(\xi) = \{D_i(\xi)\}$ of $G_\mathfrak{c}$ with respect
to $\xi$ and their 
combined sum $D^{\text{sum}}_\mathfrak{c}(\xi) = \sum D_\mathfrak{c}(\xi)$, we define 
$\mathcal{D}_\mathfrak{c}(\xi)$ as the set of the positions of the pixels of $D^{\text{sum}}_\mathfrak{c}(\xi)$ representing node activations across $G_\mathfrak{c}$. Then a concept recognition score $S_\mathfrak{c}(G_\mathfrak{c},\xi)$
is defined with a Jaccard like similarity measure similar to \citet{bau2017network}:

$$S_\mathfrak{c}(G_\mathfrak{c},\xi) = \frac{| M_\mathfrak{c}(\xi) \cap \mathcal{D}_{\mathfrak{c}}(\xi) |}{| M_\mathfrak{c}(\xi) \cup \mathcal{D}_{\mathfrak{c}}(\xi) |}$$

\begin{table}[t]
  \caption{Scene classes considered}
  \label{table:sceneclass}
  \centering
  \begin{tabular}{llllll}
    \toprule
    \textbf{Label}  & \textbf{Class Name}  & \textbf{Num. images} & \textbf{Label} & \textbf{Class Name} & \textbf{Num. images} \\
    \midrule
    0 & bathroom      & 671     &  8  & mountain snowy  & 132\\
    1 & street          & 2038  & 9   & conference room   & 168\\
    2 & office            & 112     & 10 & skyscraper       & 320\\
    3 & building facade   & 228     & 11 & corridor       & 110\\ 
    4 & airport terminal & 107    & 12 & bedroom        & 1389\\
    5 & game room     & 99    & 13 & dining room      & 412\\
    6 & living room     & 697     & 14 & highway        & 295 \\
    7 & hotel room    & 160     & 15 & kitchen          & 652\\
    \bottomrule
  \end{tabular}
\end{table}

\subsection{Recognition algorithm}
\label{subsub:greedy}
We devise a greedy algorithm to identify the $G_\mathfrak{c}$ that best 
recognizes $\mathfrak{c}$ listed as Algorithm~\ref{code:greedy}. 
The intuition behind the greedy 
approach is to find a set of feature maps that recognizes $\mathfrak{c}$ well, is as small
as possible, and is composed of feature maps that minimally `overlap', e.g. recognizes the same parts or qualities
of a concept. The latter two criteria capture the idea that a good 
distributed representation is one where the nodes of each feature map in the set 
activate over different and significant parts of the concept. Thus, in each greedy iteration, 
the algorithm searches for the feature map whose addition to $G_\mathfrak{c}$ would yield the largest improvement
in recognition score $S_\mathfrak{c}(G_\mathfrak{c},\xi)$. Large improvements would only be possible if the newly added
feature map activates over pixels representing $\mathfrak{c}$ that no other feature map in $G_\mathfrak{c}$
activates over. Moreover, this feature map cannot have significant activations over pixels that do not
represent $\mathfrak{c}$ without reducing $S_\mathfrak{c}$. Greedy iterations continue until there is no feature map whose inclusion would yield
an improvement in score greater than $\Delta$. $\Delta=0.01$ is used in the experiments below.

\begin{algorithm}[H]
\begin{algorithmic}[1]
\Procedure{greedy\_selection}{$G$, $D$, $M_\mathfrak{c}(\xi)$, $\Delta$}
\State $S_\mathfrak{c} \gets 0 $ \Comment{Score of the selected set of feature maps}
\State $G_{\mathfrak{c}} \gets \{\}$ \Comment{Set of selected feature maps}

\While{True}
  \State $tmp_s \gets 0$
  \State $g \gets \text{null}$
  \For{$k = 1$ to $|G|$}
      \State $K = G_\mathfrak{c} \cup G^k$ \Comment{Add candidate feature map $G^k \in G$ to the selected set}
        \State $D^{K}(\xi) = \sum_{k \in K} D^k(\xi)$ \Comment{Sum the deconvolutions $D^k$ of the feature maps in $K$}
        \State $S_\mathfrak{c}(K,\xi) = \frac{| M_\mathfrak{c}(\xi) \cap \mathcal{D}^{K}(\xi) |}{| M_\mathfrak{c}(\xi) \cup \mathcal{D}^{K}(\xi) |}$ \Comment{Find the new recognition score after adding $G^k$}
    \If{$S_\mathfrak{c}(K,\xi) > tmp_s$}  \Comment{Is $G^k$ better than the best candidate found so far?}
          \State $tmp_s \gets S_\mathfrak{c}(K,\xi)$ 
            \State $g \gets G^k$ 
        \EndIf
    \EndFor
    \State $G.remove(g)$  \Comment{Remove the selected feature map from $G$}
    \If{$tmp_s - S_\mathfrak{c} > \Delta$} \Comment{Does adding $g$ improve the score by more than $\Delta$?}
      \State $S_\mathfrak{c} \gets tmp_s$
        \State $G_{\mathfrak{c}}.append(g)$ \Comment{Add $g$ to the feature map set and repeat}
    \Else
      \State \Return $S_\mathfrak{c}, G_\mathfrak{c}$
    \EndIf
\EndWhile
\EndProcedure
\end{algorithmic}
\caption{Concept Localization}\label{code:greedy}
\end{algorithm}

\section{Recognition analysis}
\label{sec:recognition_analysis}
We use Algorithm~\ref{code:greedy} to recognize each concept in each given input image,
and study the relationship between its recognition quality
and a CNN's scene classification accuracy. We consider an 
AlexNet~\citep{krizhevsky2012imagenet} CNN model
trained over the Places365~\citep{zhou2016places} scene dataset and fine tune network weights using 
ADE20k~\citep{zhou2017scene}. We only consider the subset of scenes in ADE20k having at least
99 example images.
We choose this subset to ensure a sufficient number of examples are available for CNN training and 
to be able to take representative measurements of the CNN's ability to classifying a scene correctly. 
The 16 (out of the 1000+) 
scenes in ADE20k having at least 99 example images and are listed in Table~\ref{table:sceneclass}\footnote{We also omit the `misc' class of ADE20k as it is a catch-all for hard to describe scenes, even though it has over 99 images.}. 60\% of the images from each class are randomly sampled as training data during fine tuning and
40\% for testing. The fine-tuned CNN achieves a 74.9\% top-1 classification accuracy
over the testing images after 30 training epochs, which is higher than the performance of other 
CNN scene classifiers~\citep{zhou2016places}, but we note that we only test over scenes that
have an abundance of images in the ADE20K's training data. 

\begin{figure}[h]
  \centering
  \includegraphics[width=0.8\linewidth]{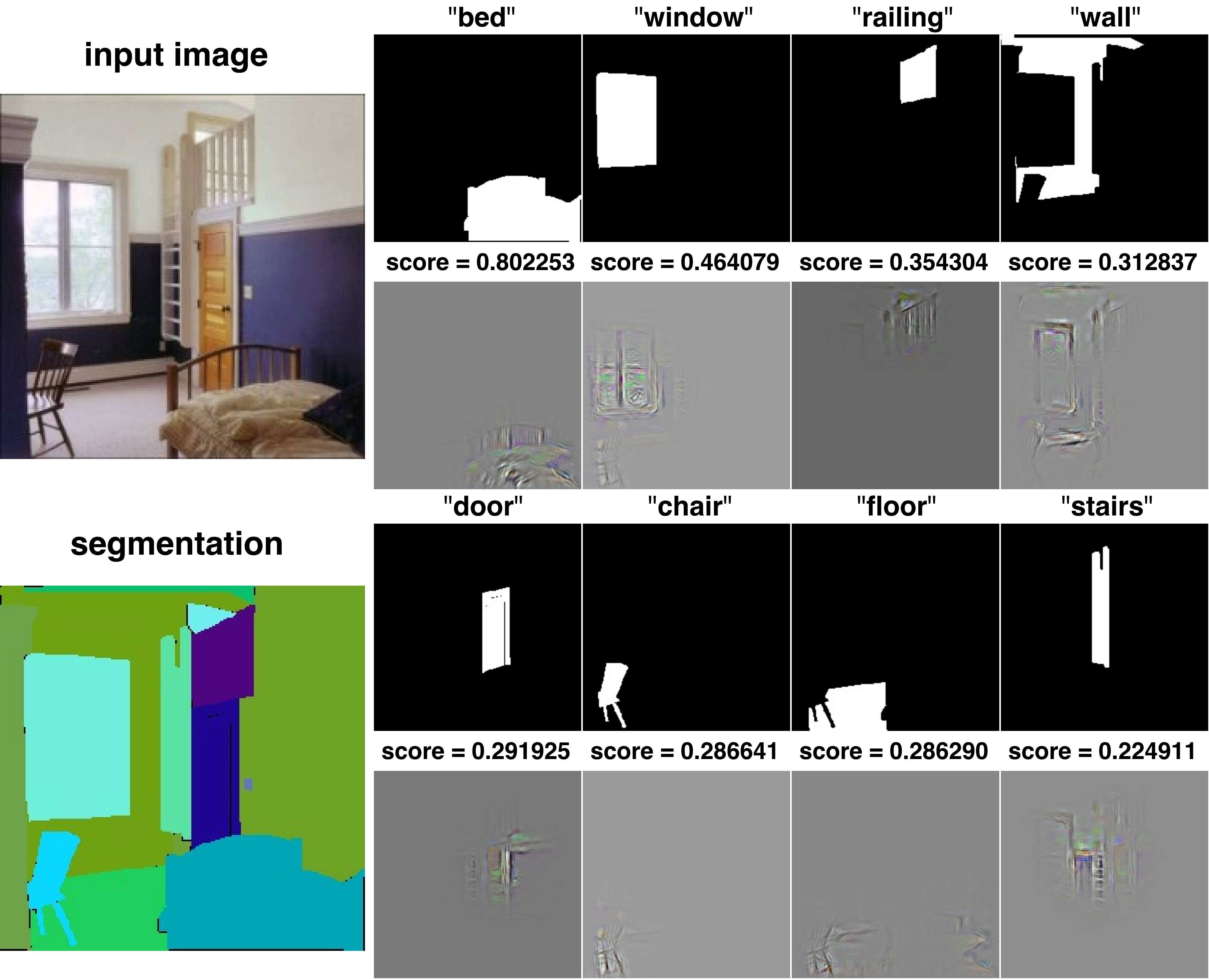}
  \caption{Concept recognition results for a given image}
  \label{fig:nips-recovery-example}
\end{figure}

\begin{figure}
\centering
\begin{minipage}{.46\textwidth}
\vspace{0px}
  \includegraphics[width=\textwidth]{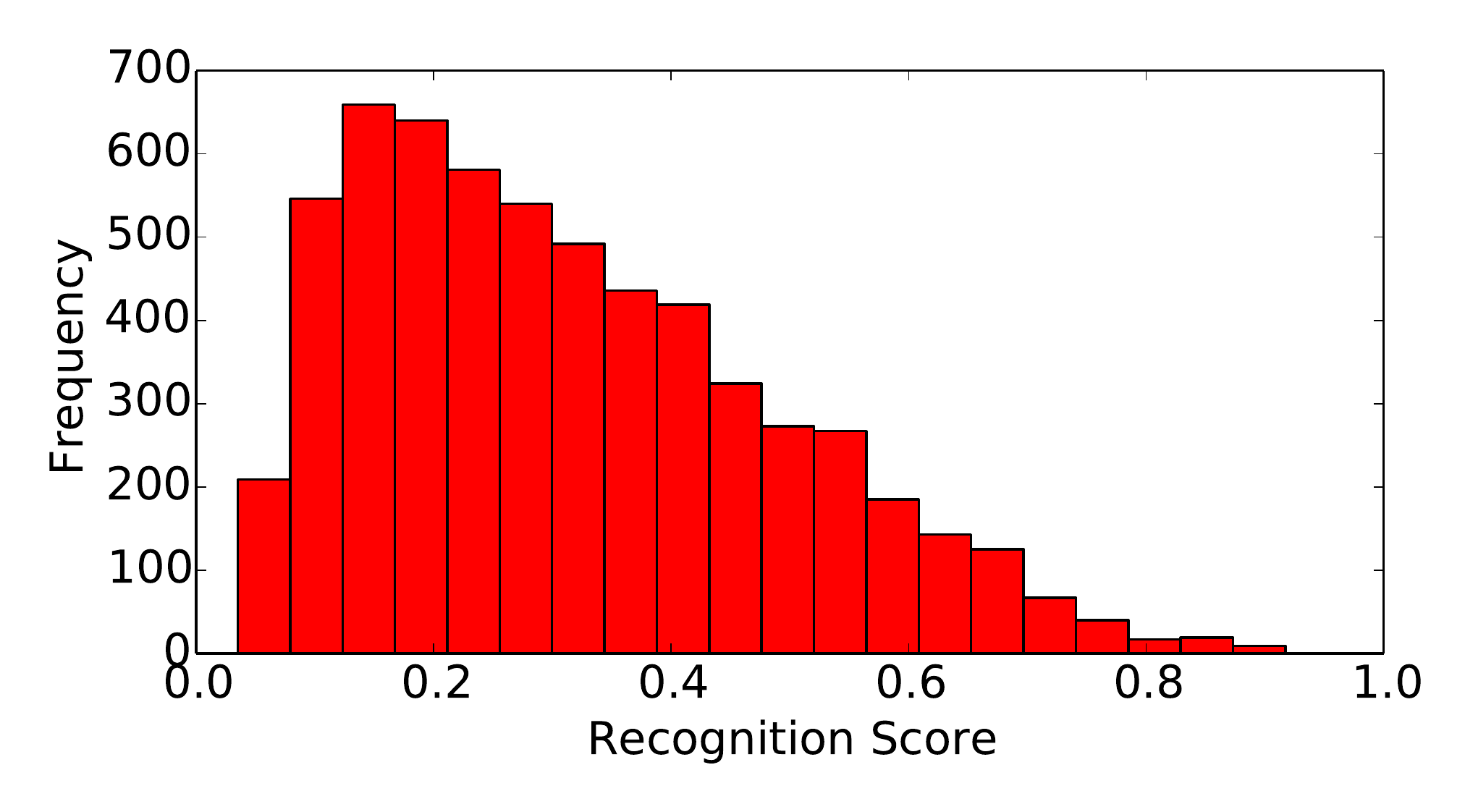}
\caption{Recognition score distribution}
\label{fig:scoredist}
\end{minipage}%
\hspace{20px}
\begin{minipage}{.48\textwidth}
\includegraphics[width=\textwidth]{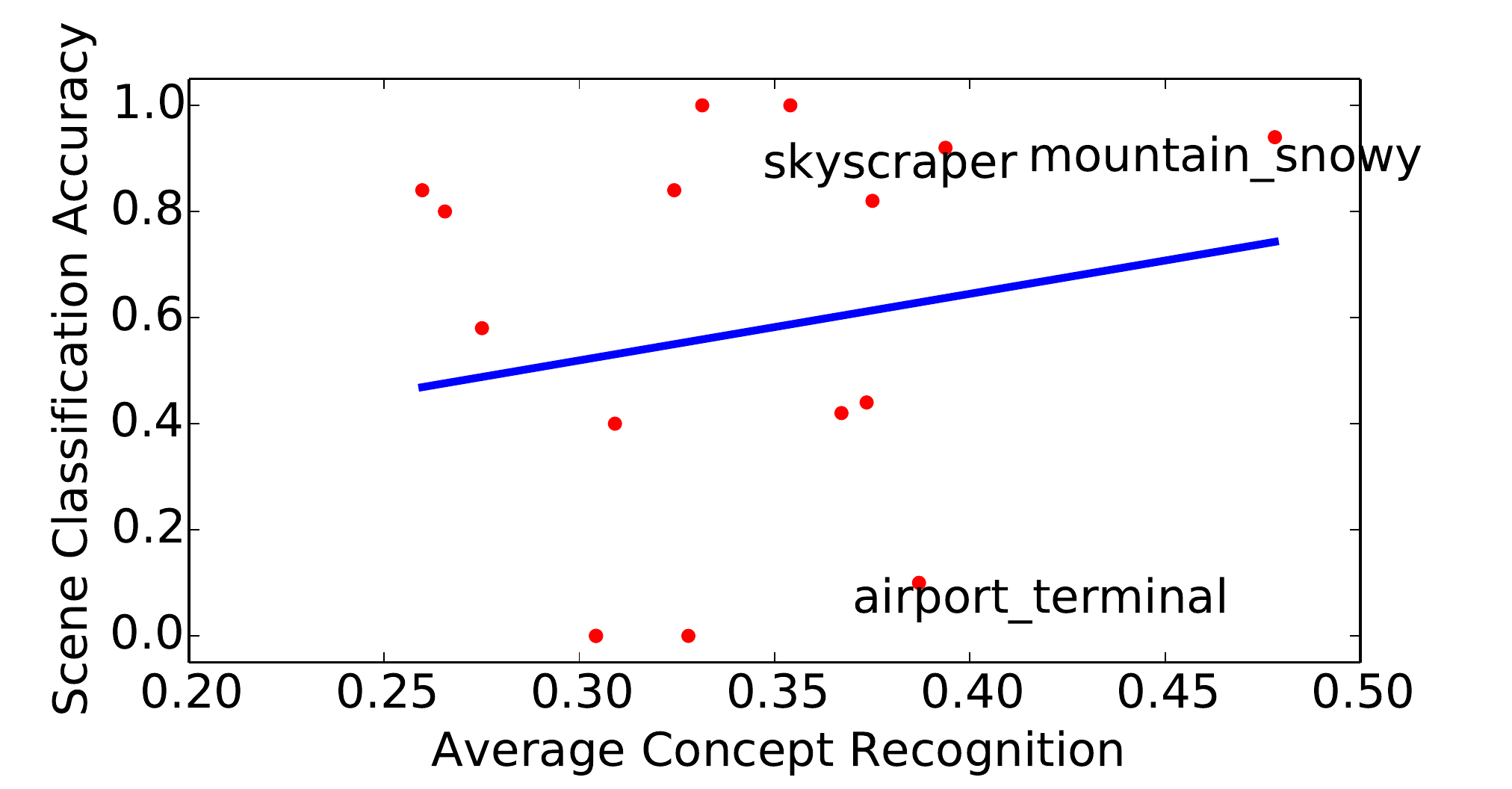}
\caption{Recognition quality vs CNN's accuracy}
\label{fig:recog}
\end{minipage}
\end{figure}

We then randomly choose 50 images from each class and compute how well their concepts are recognized by the 256 feature maps in the last convolutional layer of the CNN.
This sample of $50\times16 = 800$ images feature 370 distinct concepts. 
To get a sense of whether a recognition score is relatively ``low" or ``high", we plot the score distribution across all concepts in the sampled images in Figure~\ref{fig:scoredist}. We note that the 
mean recognition score is $0.315$ with median $0.284$, and
the lower and upper quartiles are $0.174$ and $0.429$ respectively. 
Figure~\ref{fig:nips-recovery-example} illustrates the output of Algorithm~\ref{code:greedy} in a 
sampled bedroom scene. For the eight concepts annotated in this image, the binary segmentation
mask, its label, a visualization of the sum of deconvolutions chosen by our greedy algorithm, 
and the recognition score are presented. The highest quality recognition is of the \texttt{bed}
concept, with a score ($0.802$) well above the upper quartile of the recognition score distribution 
across all concepts, a summed deconvolution that captures texture information 
about the bed and the shape and patterning of the bed frame, and activates over few pixels 
that does not represent the bed concept. The \texttt{chair} concept has a lower recognition score ($0.287$)
that happens to be close to the median of the concept recognition score distribution.
In this case, the selected feature maps are able to recognize most parts of the chair, including its legs
and back, but also happens to activate over some of the straight line and texture
patterns of the wall and floor surrounding the chair. The \texttt{stairs} concept has the lowest score ($0.225$),
caused by the feature maps' inability to activate over all pixels of the concept and also activate
across pixels representing the nearby concepts (\texttt{wall} and \texttt{door}). 

\subsection{Recognition versus performance}
\label{subsec:performance_section}
We now explore the relationship between concept recognition and CNN performance. 
For each scene and its sampled images, we compare the average recognition score of concepts within 
a scene's images against the CNN's average classification accuracy of the scene. 
Figure~\ref{fig:recog} shows only a weak linear relationship (Pearson's correlation $\rho = 0.187$), although 
there are interesting observations for some scenes. The two scenes with the best classification
and recognition scores are \texttt{skyscraper} and \texttt{mountain\_snowy}, which are scenes whose
images include concepts that are especially emblematic.
For example, the \texttt{mountain} concept is captured well across 
\texttt{mountain\_snowy} scenes ($\bar{S}^\mathtt{mountain\_snowy}_{\mathtt{mountain}} = 0.562$
where $\bar{S}^\mathfrak{s}_\mathfrak{c}$ denotes the average recognition of concept $\mathfrak{c}$ across
the sampled scenes of $\mathfrak{s}$) 
and concepts like \texttt{skyscraper}, \texttt{sky}, and \texttt{building} are identified well in 
\texttt{skyscraper} scenes ($\bar{S}^{\texttt{skyscraper}}_{\mathtt{sky}} = 0.532$, $\bar{S}^\texttt{skyscraper}_{\mathtt{building}}=0.362$, $\bar{S}^\texttt{skyscraper}_{\mathtt{skyscraper}}=0.407$).
\texttt{airport\_terminal} is a challenging scene for the CNN to identify despite achieving
high average concept recognition. This may be due to strong recognitions for concepts like 
\texttt{floor} and \texttt{ceiling} ($\bar{S}^\texttt{airport\_terminal}_{\mathtt{floor}}=0.585$, 
$\bar{S}^\texttt{airport\_terminal}_{\mathtt{ceiling}}=0.559$) that appear in at least 45 of the 50 sampled \texttt{airport\_terminal} images, but these concepts are generic and could apply to any kind of indoor scene. Concepts better capturing the notion of an airport terminal are also recognized, e.g., \texttt{armchair} ($\bar{S}^\texttt{airport\_terminal}_{\mathtt{armchair}} = 0.555$) and \texttt{shops} ($\bar{S}^\texttt{airport\_terminal}_{\mathtt{shops}} = 0.548$), but they emerge in only one of the sampled images.

\subsection{Sparse concepts}
The \texttt{airport\_terminal} example suggests that there may be 
particular types of concepts that have stronger or weaker relationships to a CNN's decisions. 
We first consider `sparse' concepts, which are concepts
appearing in a small number of images within a scene 
(we quantify this notion with a \textit{popularity} score in the sequel). 
Sparse concepts may not appear often enough
during training for a CNN to learn to recognize well or to relate with a particular scene. 
For example, while the CNN is able to recognize the $\texttt{armchair}$ and $\texttt{shops}$ 
concepts in an \texttt{airport\_terminal} well, their infrequency 
could mean the CNN does not have enough observations to establish a relationship between
these concepts and the scene label. 

\begin{figure}[h]
  \includegraphics[scale=0.279]{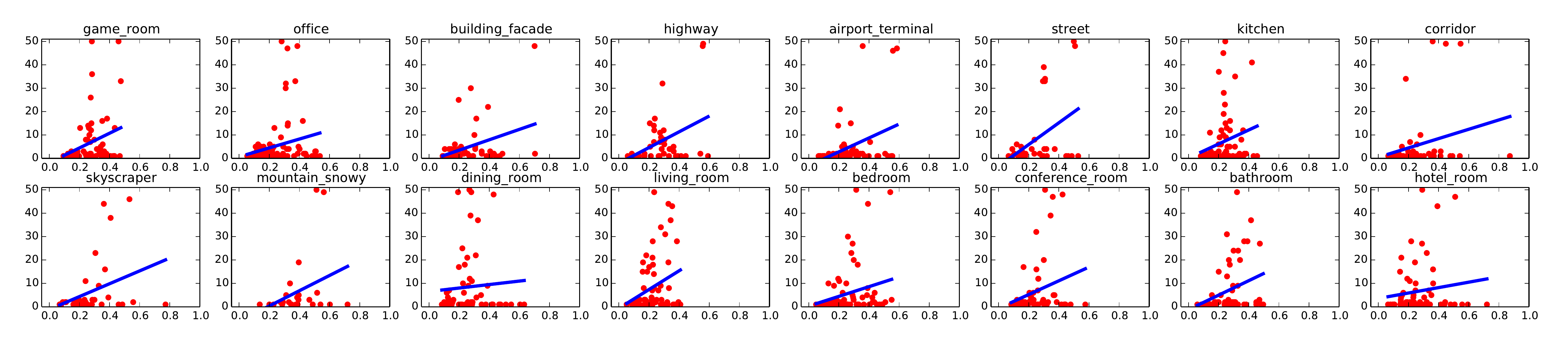}
   \caption{Average concept recognition (x-axis) vs. number of concept occurrences 
  (y-axis) per scene }
  \label{fig:recpop}
\end{figure}

Figure~\ref{fig:recpop} explores the prevalence of concepts and how well they are recognized 
across each of the 16 scene classes. 
It illustrates that, for every class, there are a majority of concepts that emerge in less
than 10 of the 50 images sampled from each scene. Scenes that are relatively uniform in the way they look,
for instance \texttt{skyscraper}, \texttt{mountain\_snowy}, and \texttt{street} scene, have fewer 
sparse concepts. Moreover, such scenes tend to have their non-sparse
concepts recognized strongly by the CNN (reflected by the steeper 
slopes of the linear fits in their scatter plots). Scenes that are non-uniform in what they could look like,
for example \texttt{bedroom}, \texttt{hotel\_room}, and \texttt{dining\_room} images
that depict different styles and design, tend to exhibit a larger number of sparse concepts. 
But some of these sparse concepts have 
high recognition scores (resulting in shallower slopes of the linear fits in their scatter plots), suggesting 
that the CNN learns to recognize them. This may be because 
a sparse concept could be observed across a large number of different scenes. For example,
although not every \texttt{bedroom} has a \texttt{chair}, one can imagine a \texttt{chair} 
to appear across a variety of different scenes, giving a CNN enough examples to learn to recognize 
this concept. 

The figure and discussion suggest the following hypothesis:
the fewer the number of sparse concepts present and the 
greater the number of well recognized non-sparse concepts appear across the images of a scene, the higher the chance is that the CNN can correctly identify the scene. Moreover, scenes whose images are 
dominated by a variety of sparse concepts should prove to be more challenging for the CNN to classify.
To test this, we plot 
the slope of the linear fit of each scatter plot from Figure~\ref{fig:recpop} against the CNN's 
accuracy for each scene in Figure~\ref{fig:slopeacc}. The
moderate linear relationship (Pearson's $\rho = 0.444$) suggests that many non-sparse, well recognized
concepts are associated with correct CNN decisions, lending support for the hypothesis.

\subsection{Unique and misleading concepts}
We now investigate non-sparse concepts further. Intuitively, non-sparse concepts may have greater
benefit to correct CNN decisions if they appear across a smaller
number of different types of scenes. For example, concepts
like \texttt{sand} and \texttt{shell} may be present 
in many beach scenes, are closely associated with the notion of beach, and are unlikely to 
appear in other types of scenes. Thus, 
high quality recognition of \texttt{sand} and \texttt{shell}
concepts would help a CNN to classify \texttt{beach} scenes correctly. On the other hand, 
non-sparse concepts emerging across a variety of scenes may be less helpful. For example, 
since we expect most images of indoor scenes 
to include concepts like \texttt{wall}, \texttt{floor}, or \texttt{ceiling}, their recognition 
may not help a CNN differentiate between different indoor scenes. In fact, these recognitions
may be of limited help in the best case and could confuse or mislead a CNN to make a wrong 
classification in the worst case. 

\begin{figure}
\centering
\begin{minipage}{.46\textwidth}
\vspace{6px}
  \includegraphics[width=\textwidth]{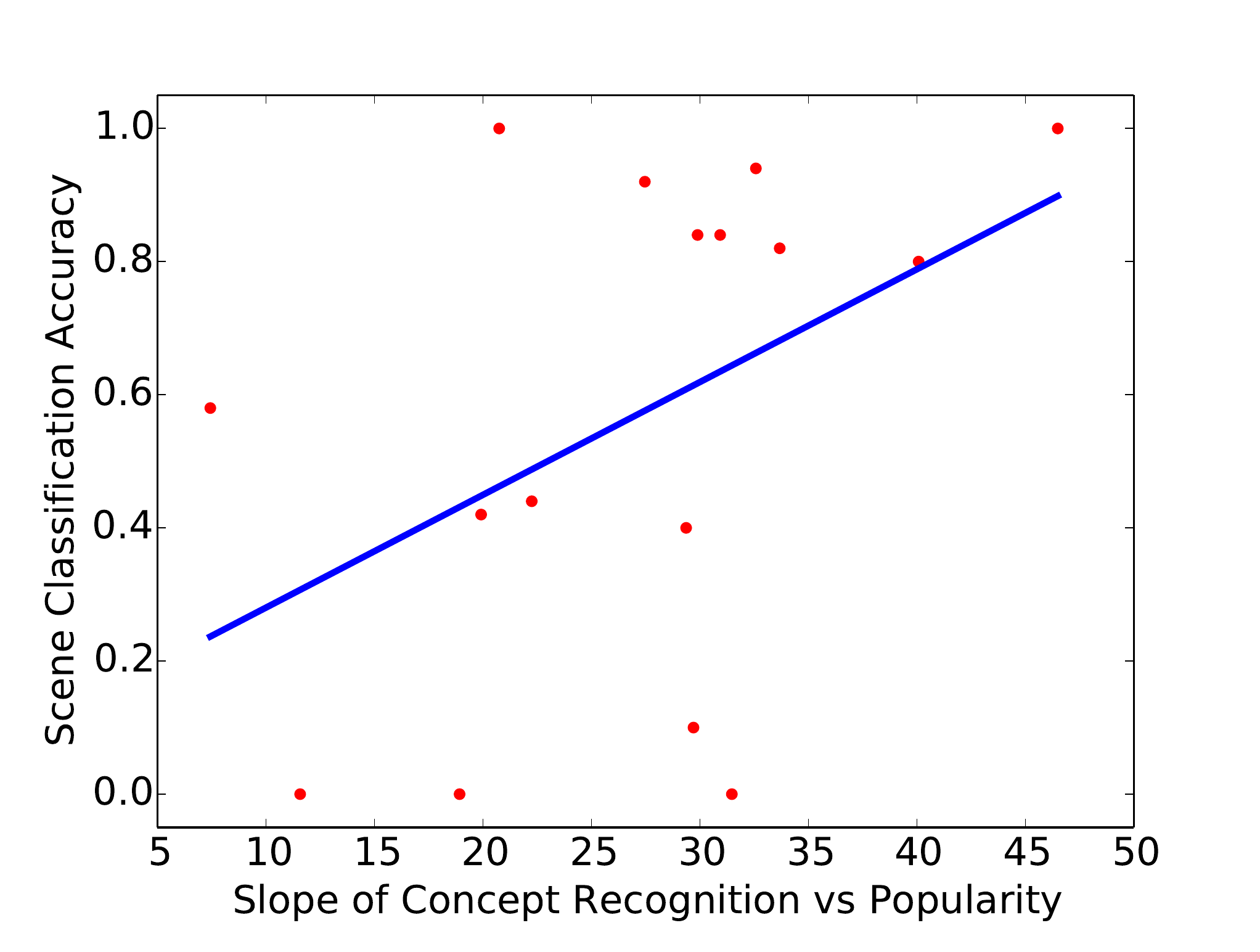}
\caption{Slope of sparse concept recognition (Figure~\ref{fig:recpop}) vs CNN's accuracy}
\label{fig:slopeacc}
\end{minipage}%
\hspace{20px}
\begin{minipage}{.48\textwidth}
\includegraphics[width=\textwidth]{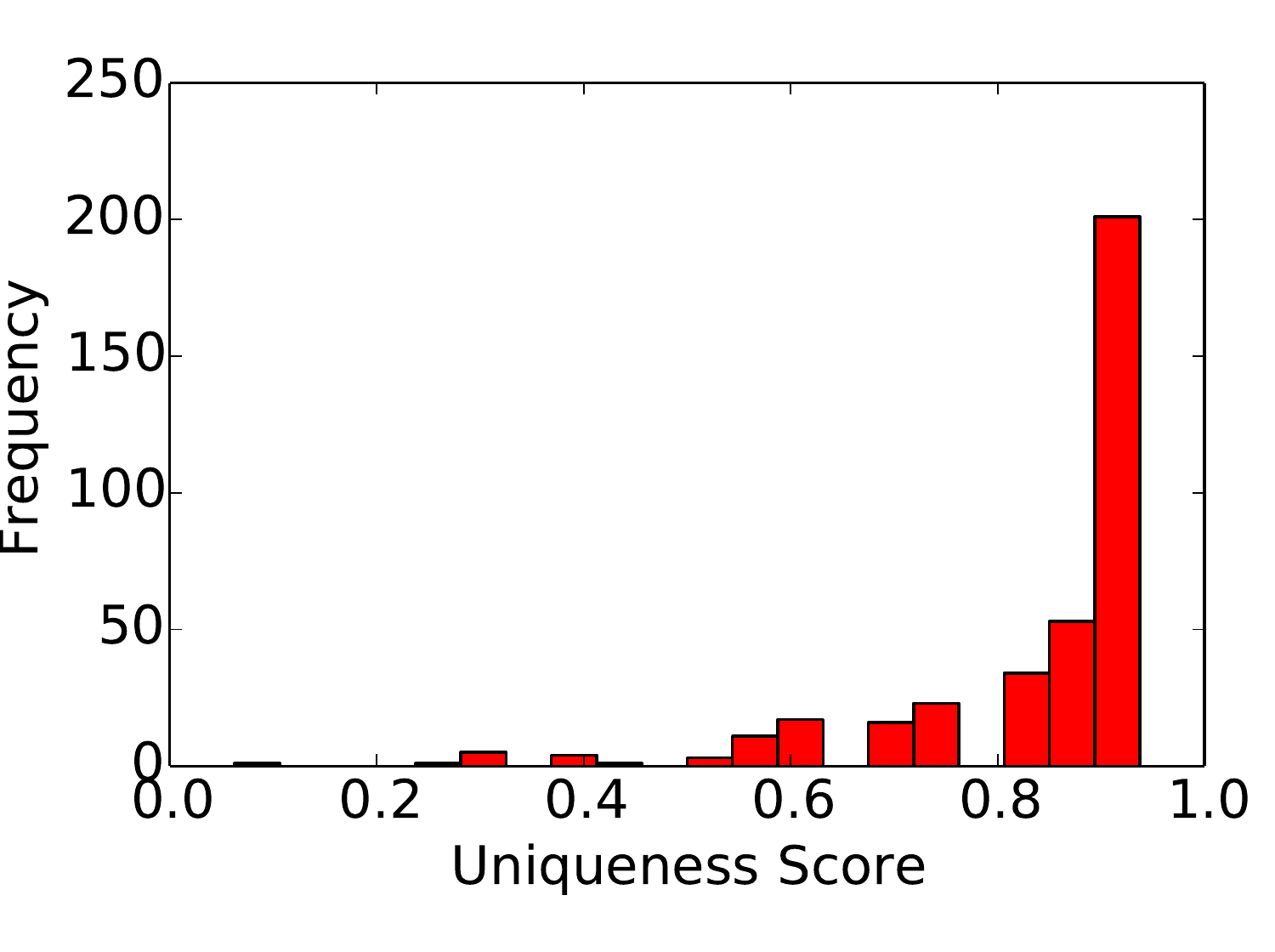}
\caption{Uniqueness score distribution}
\label{fig:uniquedist}
\end{minipage}
\end{figure}

To explore these ideas, we compute a {\em uniqueness} score of a concept that 
reflects the variety of scenes it appears in. 
The uniqueness $U(\mathfrak{c})$ of a concept $\mathfrak{c}$ is calculated as:
 
$$U(\mathfrak{c}) = 1-\frac{\text{\# of scene classes $\mathfrak{c}$ appears}}{\text{\# of scene classes}}$$
Figure~\ref{fig:uniquedist} gives the distribution of the uniqueness scores of each concept.
It is skewed, with its average uniqueness score at $0.845$, and its lower quartile, median, and upper quartile
is $0.8125$, $0.9375$, and $0.9375$ respectively.
210 of the 370 concepts appear in only one scene class, although many of these concepts are likely to be sparse. Following the fact that many
of the scenes used in our analysis (listed in Table~\ref{table:sceneclass}) are indoors, concepts
with the least unique scores pertain to generic aspects of a room. For example, the 
concepts having the three lowest uniqueness scores are $U(\mathtt{wall}) = 0.063$, 
$U(\mathtt{floor}) = 0.25$, and $U(\mathtt{door}) = U(\mathtt{plant}) = U(\mathtt{window}) = U(\mathtt{ceiling}) = U(\mathtt{picture}) = 0.3125$. 

We hypothesize that the recognition of unique concepts helps a CNN make correct classifications, 
and that concepts with low uniqueness scores may `mislead' a CNN. We evaluate this hypothesis
by comparing the CNN's classification accuracy to the average recognition score calculated on ``unique" concepts and ``misleading'' concepts respectively. A concept $\mathfrak{c}$ is labeled as ``unique" if its uniqueness score $U(\mathfrak{c}) > \alpha$ for a uniqueness threshold $\alpha$. 
However, we recall from Figure~\ref{fig:recpop} that a number
of unique concepts are likely to be `sparse', thus hindering classification accuracy (Figure~\ref{fig:slopeacc}). We thus filter away sparse concepts by defining a {\em popularity} score
$P(\mathfrak{c})$ with respect to some scene by: 

$$P(\mathfrak{c}) = \frac{\text{\# of images $\mathfrak{c}$ appears in a scene class}}{\text{\# of images sampled from a scene class}}$$
and only consider concepts whose $P(\mathfrak{c}) > \beta$ for a popularity threshold $\beta$. 

\begin{figure}[h]
  \centering
  \includegraphics[width=1\linewidth]{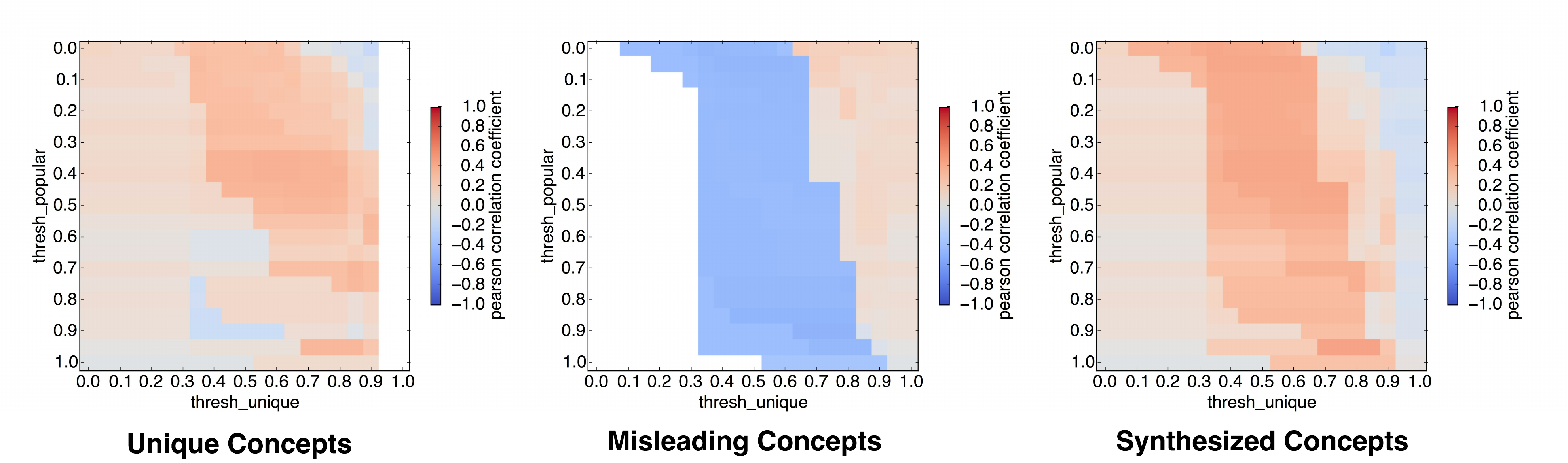}
  \caption{Heatmap for PCC calculated upon ``unique'' concept, ``misleading'' concept, and ``synthesized'' of unique and misleading concepts using different thresholds.}
  \label{fig:nips-heatmap}
\end{figure}

We then compute Pearson's correlation coefficient $\rho$ between the CNN's accuracy over each scene class against the 
average recognition score on ``unique'' and ``misleading'' concepts respectively for various values of $\alpha$ and $\beta$. 
Figure~\ref{fig:nips-heatmap} presents $\rho$ over a grid of the two thresholds, varying their values in
increments of $0.05$ between $0$ and $1$. 
The left heatmap shows $\rho$ when only unique concepts are considered.
Most of the area shows a positive relationship between the unique concepts recognition quality and CNN accuracy. Larger uniqueness and popularity thresholds $\alpha$ and $\beta$, making the set of unique concepts even
smaller, lead to an even stronger relationship. 
Note that there is no concept having $U(\mathfrak{c})\geq 0.95$, causing empty cells in the right most two 
columns.
The middle heatmap only considers misleading concepts. The shaded blue areas indicate a negative relationship between the misleading concepts recognition quality and the model performance.
For most valid settings of $\beta$, when $U(\mathfrak{c}) < 0.7$, there exists a moderate strong negative correlation. This provides some evidence
that the recognition of misleading concepts, e.g. those 
concepts appearing across many different scene types, may be hindering a CNN's ability to classify scenes correctly. 
The right heatmap reports $\rho$ using a ``synthesized'' 
average concept recognition score, which is defined for each scene class by 
$S_\text{syn} = \nicefrac{(S_\text{unique} + 1.0 - S_\text{mislead})}{2}$ where $S_\text{unique}$
is the average concept recognition score over the unique concepts and $S_\text{mislead}$
is the same but over misleading concepts. This synthetic score 
unifies the results from the unique and misleading heatmaps together in search of 
threshold settings that maximize $\rho$ over unique concepts and minimize $\rho$ over 
misleading concepts. We find the highest positive correlation of $\rho = 0.521$ using the synthetic scores
when $\beta=0.4$ and $\alpha=0.55$. 
At these thresholds, we find $\rho = 0.454; (p = 0.078)$ over the unique concepts and $\rho = -0.528; (p = 0.036)$
on the misleading concepts. The $p$-values for these correlation scores, computed over $n=16$
classes, indicate a significant negative correlation between misleading concept recognition and
CNN's accuracy, and a moderate positive correlation between unique concept recognition and CNN's accuracy. 

\section{Conclusions and future work}
\label{sec:conclusions}

This paper investigated the relationship between a CNN's recognition of input concepts and classification
accuracy. A novel approach was developed to quantify how well a concept (specifically, an object in an image) is recognized across the latest convolutional layer of a CNN. Analysis using image object
annotations in the ADE20k scene dataset revealed a weak relationship between the average 
recognition of image concepts in a scene and classification accuracy. We found evidence to suggest 
that the relationship is hindered by recognized concepts that are ``sparse'', or appear in a small number
of images of a scene and by ``misleading'' concepts that appear in many images across
many different scenes. Recognizing ``unique'' concepts, which appear often but in a limited set of scenes, 
is moderately positively correlated with the CNN's classification accuracy. 

For future work, we will analyze which feature maps are necessary to accurately model each object in the scene.
The effects of ``unique'', ``misleading'', and ``sparse'' concepts will be explored in more detail. 
In particular, we will investigate common misclassifications for a 
scene and seek explanations by the recognized concepts that are (not) common
between them.
We will study the effect of ``sparse'' concepts on CNN classification via their
occlusion in an image. We will also explore the mechanics of how concept recognitions 
impact downstream network activations leading to a decision and devise a measure of the 
importance of concept recognition to CNN decision making.

\bibliographystyle{natbib.bst}
\bibliography{HCBD_ref}

\end{document}